\documentclass[conference]{IEEEtran}
\IEEEoverridecommandlockouts

\usepackage{cite}
\usepackage{amsmath,amssymb,amsfonts}
\usepackage{algorithmic}
\usepackage{graphicx}
\usepackage{textcomp}
\usepackage{xcolor}
\usepackage{lipsum}
\usepackage{booktabs}
\usepackage{array}
\usepackage{fancyhdr}
\usepackage{amsmath}
\usepackage{float}
\usepackage{xcolor}
\usepackage{hyperref}
\hypersetup{
    colorlinks=true,
    linkcolor=blue,
    filecolor=blue,
    urlcolor=blue,
    citecolor=blue,
}

\def\BibTeX{{\rm B\kern-.05em{\sc i\kern-.025em b}\kern-.08em
    T\kern-.1667em\lower.7ex\hbox{E}\kern-.125emX}}
\renewcommand{\thefigure}{\arabic{figure}}

\fancypagestyle{plain}{
    \fancyhf{}
    \rfoot{\thepage}
    
}
\begin{document}

\title{Multispectral Remote Sensing for Weed Detection in West Australian Agricultural Lands}

\vspace{-6mm}
\author{
\IEEEauthorblockN{Haitian Wang\(^1\), Muhammad Ibrahim\(^2\),Yumeng Miao\(^1\), Dustin Severtson\(^1\,^2\),Atif Mansoor\(^1\), Ajmal S. Mian\(^1\)}
\IEEEauthorblockA{
\(^1\)The University of Western Australia, \(^2\)Department of Primary Industries and Regional Development
}
\vspace{-2mm}
\thanks{979-8-3503-7903-7/24/\$31.00 ©2024 IEEE}
}

\maketitle

\begin{abstract}
The Kondinin region in Western Australia faces significant agricultural challenges due to pervasive weed infestations, causing economic losses and ecological impacts. This study constructs a tailored multispectral remote sensing dataset and an end-to-end framework for weed detection to advance precision agriculture practices. Unmanned aerial vehicles were used to collect raw multispectral data from two experimental areas (E2 and E8) over four years, covering 0.6046 km² and ground truth annotations were created with GPS-enabled vehicles to manually label weeds and crops. The dataset is specifically designed for agricultural applications in Western Australia. We propose an end-to-end framework for weed detection that includes extensive preprocessing steps, such as denoising, radiometric calibration, image alignment, orthorectification, and stitching. The proposed method combines vegetation indices (NDVI, GNDVI, EVI, SAVI, MSAVI) with multispectral channels to form classification features, and employs several deep learning models to identify weeds based on the input features. Among these models, ResNet achieves the highest performance, with a weed detection accuracy of 0.9213, an F1-Score of 0.8735, an mIOU of 0.7888, and an mDC of 0.8865, validating the efficacy of the dataset and the proposed weed detection method.
\end{abstract}

\begin{IEEEkeywords}
Weed detection, Remote Sensing, Multispectral Image, Vegetation Indices. 
\end{IEEEkeywords}

\begin{figure*}[htbp]
\centering
\includegraphics[width=\textwidth]{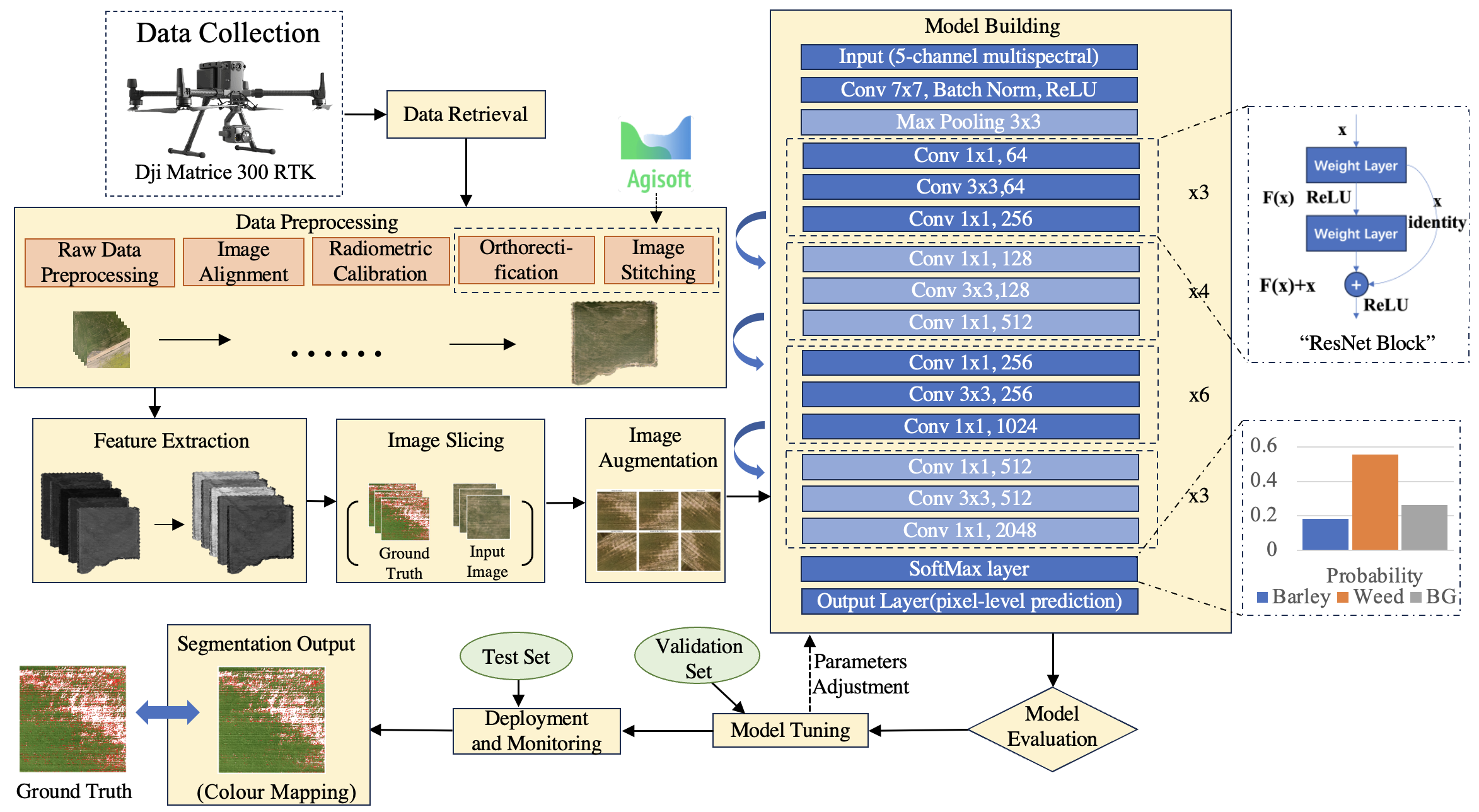} 
\vspace{-8mm}
\caption{\footnotesize{Workflow of our proposed end-to-end weed detection framework with a ResNet-50 model: The workflow starts with data retrieval, preprocessing, feature selection, and image slicing. The sliced sub-images are divided into three groups: training set, validation set, and test set. The training set is augmented and used to train a ResNet-50 model. The validation set is used to determine the optimal hyperparameters. These optimal hyperparameters are applied to train the optimized model, which is then tested and analyzed with the test set.}}
\label{pipeline}
\vspace{-6mm}
\end{figure*}

\section{Introduction}
\textcolor{black}
Western Australia faces a severe agricultural crisis due to extensive weeds. The annual expenditure on weed control is substantial, reflecting the significant burden weeds place on agricultural production. The cropping industries alone incur a staggering cost of \$1,206 million each year\cite{hafi2022cost}. To address this urgent issue, after extensive research, we have identified that long-term reliance on chemical herbicides has led to significant resistance in weeds like ryegrass and barley grass, causing them to proliferate rapidly \cite{beckie2019herbicide}. Confronted with these challenges, this paper aims to adopt precision agriculture methods which leverage deep learning models in computer vision to achieve precise identification and classification of weeds in remote sensing images, thereby improving management efficiency and reducing environmental impact \cite{karunathilake2023path}.

\textcolor{black}{Research efforts towards addressing the challenges posed by weeds through precision agriculture has shown that the distribution of weeds exhibits certain spatial and temporal variability, which necessitates the use of remote sensing technology to capture these \cite{sa2018weednet}. Remote sensing, particularly multispectral imaging, has been widely adopted in agriculture due to its capability to provide high spectral resolution imagery \cite{sishodia2020applications}. One of the key advantages of remote sensing is its ability to cover extensive agricultural areas with ease, making it indispensable for large-scale monitoring that ground based methods cannot achieve \cite{khanal2020remote}.}

\textcolor{black}{The unique geographical conditions of Western Australia, including soil types, climate conditions, and sunlight intensity, pose significant challenges in directly applying methods developed using datasets from other regions such as the United States and Europe to our local context \cite{hassani2020predicting}. Many external datasets, such as those from Google Earth and satellite imagery, often fall short in resolution and timeliness, making them less suitable for the precision agriculture needs of the WA where up-to-date and detailed data are crucial. Therefore, to achieve accurate identification of weeds using remote sensing and computer vision, the primary task is to collect and construct a multispectral agricultural geoinformed dataset tailored to local conditions.}

\textcolor{black}{A major contributions of this paper is the development of the first multispectral remote sensing dataset specifically designed for agricultural applications in Western Australia. 
Our data collection spanned four years (2020 to 2023) to develop a multispectral image dataset specifically tailored for agricultural geoinformation in Western Australia. To ensure the dataset's accuracy and utility, we employed advanced unmanned aerial vehicle (UAV) technologies, including the DJI Matrice 300 RTK and the DJI P4 Multispectral (P4M)\cite{dji2019p4} remotely piloted aircraft system (RPAS), to capture high-resolution agricultural imagery over the extensive farmlands in the Kondinin region. Utilizing these UAV technologies, raw multispectral data were collected from two experimental areas marked as E2 and E8, covering 0.6046 km² and ground truth annotations were created by using GPS-enabled vehicles to manually label weeds and crops.}

\textcolor{black}{Another major contribution of this paper is the development of an end-to-end framework for weed detection, as shown in Fig.~\ref{pipeline}. The proposed framework effectively encompasses raw image preprocessing, feature selection, deep neural network model training, and compilation of prediction results. The raw multispectral images underwent a series of preprocessing steps, including image denoising, image alignment, radiometric calibration, and image stitching. Subsequently, through an in-depth analysis of relevant literature on remote sensing and agricultural technologies, as well as a detailed examination of the dataset, we selected several valuable vegetation indices (NDVI, GNDVI, EVI, SAVI, MSAVI) as key features. Finally, we utilized multiple deep learning techniques using models such as ResNet\cite{bah2018deep}, U-Net\cite{ronneberger2015unet}, DeepLabV3+\cite{chen2018deeplab}, InceptionV3\cite{olsen2019deepweeds}, and SegNet\cite{badrinarayanan2017segnet}, trained for weed detection on the constructed multispectral dataset and selected key features. Among these, the ResNet-50 model achieved the highest performance metrics, with an accuracy of 0.9213, an F1-Score of 0.8735, an mIOU of 0.7888, and an mDC of 0.8865, validating the efficacy of the dataset and our end-to-end framework. The proposed framework not only establishes a solid baseline for future research but also provides a practical example of the dataset's application, demonstrating its efficacy and versatility in addressing real-world agricultural challenges.}

\textcolor{black}{Section~\ref{sec:Related work} reviews related work, providing an overview of existing research and methodologies in weed and crop classification using multispectral data and deep learning models. Section~\ref{sec: Methodology} details the preprocessing methods employed to construct our high-resolution multispectral dataset and outlines our strategy for feature selection. Section~\ref{sec:Data Collection} describes the method for UAV setting and raw data collection, including flight path planning and calibration procedures. Section~\ref{sec:Experiment} explains the experimental setup and process, encompassing dataset formation, model evaluation metrics and result analysis. Section~\ref{sec:Conclusion} concludes the paper by summarizing the key findings and proposing directions for future research in precision agriculture and remote sensing.}

\vspace{-1mm}
\section{Related Work}
\vspace{-1mm}
\label{sec:Related work}
\textcolor{black}
In the domain of precision agriculture, several studies have employed deep learning models for weed and crop classification using multispectral data. This section reviews relevant literature that informs our approach, focusing on how previous researchers have tackled similar problems.

\textcolor{black}Bah et al.  \cite{bah2018deep} used ResNet for weed detection in polyhouse-grown bell peppers. They demonstrated that ResNet's deep architecture and residual connections can effectively handle the complex variability in agricultural imagery, resulting in high classification accuracy. This work supports our choice of ResNet for its robustness in handling complex and high-dimensional multispectral data. \textcolor{black}Ronneberger et al. \cite{ronneberger2015unet} introduced U-Net for image segmentation, which has been successfully adapted for agricultural applications\cite{mendoza2022cnn}. U-Net's encoder-decoder structure and its capability to perform precise pixel-level segmentation make it an ideal choice for distinguishing between crops and weeds in high-resolution multispectral images.

\textcolor{black}Chen et al.  \cite{chen2018deeplab} applied the DeepLab model for semantic segmentation in agricultural fields. DeepLab's use of atrous convolution and Conditional Random Fields (CRFs) enhances boundary delineation and maintains high-resolution features\cite{chen2018deeplab}, which is critical for accurately segmenting the intricate patterns of weeds and crops. \textcolor{black}Badrinarayanan et al. \cite{badrinarayanan2017segnet} developed SegNet for efficient semantic segmentation, demonstrating its effectiveness in processing high-resolution remote sensing data. SegNet’s encoder-decoder framework is particularly suited for applications requiring detailed segmentation of agricultural fields, where computational efficiency is also a priority .

\textcolor{black}The work by Szegedy et al. \cite{szegedy2016rethinking} on InceptionV3 showed how factorized convolutions and diverse filter sizes could handle large and varied datasets. This makes InceptionV3 a strong candidate for our study, given its ability to manage the diverse spectral bands in multispectral agricultural data and provide accurate classification results.

\textcolor{black}Recent advancements have focused on unsupervised domain adaptation (UDA) techniques to improve model performance across different fields. Huang et al. \cite{huang2024unsupervised} proposed an unsupervised domain adaptation framework using greedy pseudo-labeling, which optimizes pseudo-label selection to enhance weed segmentation under varied conditions . This approach mitigates overfitting by monitoring covariance during co-training, ensuring robust model adaptation across different agricultural contexts.
\textcolor{black}In another study, an unsupervised classification algorithm was developed for early weed detection in row-crops by combining spatial and spectral information. This method leveraged Fourier Transform for row orientation detection and NDVI(Normalized Difference Vegetation Index, an index that quantifies vegetation by measuring the difference between near-infrared and red light reflectance.) for vegetation discrimination, improving classification results by integrating spatial and spectral data \cite{mendoza2022cnn}.

\textcolor{black}These studies collectively highlight the effectiveness of advanced deep learning models in the context of weed and crop classification using multispectral datasets. They provide a solid foundation for our approach, validating our model choices and guiding our methodological framework.

\vspace{-1mm}
\section{Methodology}
\label{sec: Methodology}

This section provides an overview of the end-to-end framework employed in our study to achieve precise weed detection using UAV-based multispectral imaging and deep learning techniques. Our methodology encompasses several critical stages including data preprocessing, feature selection, and weed detection. Starting with raw data collection from UAVs, we preprocess the images through denoising, radiometric calibration, and alignment. Then, stitching is applied to ensure a high-quality map covering the entire observation area is build. Feature selection is then performed to extract relevant vegetation indices that enhance the classification process. Finally, multiple deep learning models are trained and evaluated to identify the most effective model for weed detection.

\subsection{Data Preprocessing}
\label{sec: Data Preprocessing}
Data preprocessing includes the preprocessing of raw images, their alignment, radiometric calibration, orthorectification and image stitching. 

\subsubsection{Preprocessing of Raw Images}
Preprocessing primarily involves denoising to enhance image quality by eliminating noise introduced by sensors and environmental factors. In this project, the primary sources of noise include high-frequency noise and impulse noise. High-frequency noise, typically caused by sensors or circuitry, appears as small, random specks in the image. Impulse noise manifests as isolated bright or dark spots, usually resulting from sensor malfunctions or data transmission errors. Additionally, environmental noise, such as fluctuations in light source stability and electromagnetic interference from nearby electronic devices, can further degrade image quality. To ensure the accuracy of subsequent analysis, a median filter is applied for denoising \cite{hwang1995adaptive}. 
The median filter is particularly effective at removing impulse noise while preserving edge information in the image, thus maintaining the integrity of spectral data. 

\subsubsection{Image Alignment}
\textcolor{black}
{To ensure pixels of each band correspond precisely to the same location, two main steps are employed to achieve image alignment: feature point matching and image registration.}

\begin{figure}
\includegraphics[width=0.470\textwidth]{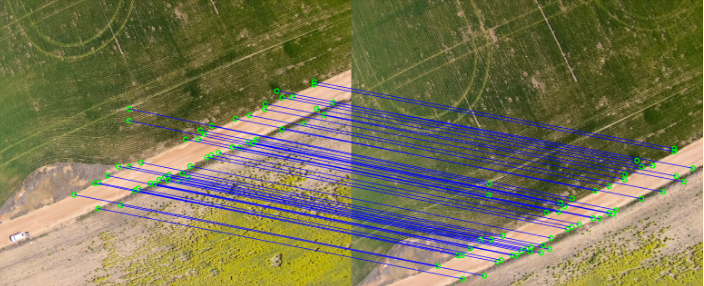} 
\vspace{-3mm}
  \caption{\footnotesize{SIFT feature detection for image alignment. Keypoints are shown on green and the correlations are shown in blue.}}
\label{sift}   
\vspace{-6mm}  
\end{figure}

{Firstly, the Scale-Invariant Feature Transform (SIFT) algorithm \cite{ghazal2024cv} is used to detect feature points in each image band as it is shown in Fig.~\ref{sift}. SIFT identifies key points in the images and computes descriptors for these points, which are invariant to scale and rotation \cite{ghazal2024cv}. The distances between feature point descriptors in different band images are calculated to find matching feature point pairs. The nearest neighbor algorithm is employed for feature point matching, and the Random Sample Consensus (RANSAC) algorithm is applied to eliminate erroneous matches, retaining only reliable feature point pairs. Using the matched feature point pairs, the affine transformation matrix between the images is computed, and this matrix is applied to transform each band image, aligning them to the same coordinate system \cite{lowe2004distinctive}. Then, image registration based on correlation coefficients and mutual information is performed \cite{zitova2003image}.}

\subsubsection{Radiometric Calibration}
To ensures consistent radiance across spectral bands captured at different times, a reference target calibration method is used, utilizing a reference target with known reflectance to calibrate and eliminate radiance discrepancies between different bands \cite{hwang1995adaptive}. 
Firstly, the brightness values of each band are normalized. This involves computing the mean ($\mu$) and standard deviation ($\sigma$) of the original image brightness values ($I_{original}$) and normalizing them according to the following equation:

\begin{equation}
\label{eq:ndvi1}
I_{normalized} = \frac{I_{original} - \mu}{\sigma},
\end{equation}
where $I_{normalized}$ represents the normalized image brightness.

Next, before each flight mission, a reference target with known reflectance is captured using the multispectral camera under consistent lighting conditions. The environmental parameters, such as illumination conditions and solar elevation angle, are recorded. The reflectance calibration factor ($C_{calibration}$) for each band is then calculated using the known reflective value ($R_{target}$) of the reference target and the measured brightness value ($I_{measured}$) as follows:

\begin{equation}
\label{eq:ndvi2}
C_{calibration} = \frac{R_{target}}{I_{measured}}.
\end{equation}

Finally, these calibration factors are applied to calibrate the images of all bands using the following equation:

\begin{equation}
\label{eq:ndvi3}
I_{calibrated} = I_{measured} \times C_{calibration},
\end{equation}
where $I_{calibrated}$ represents the calibrated image brightness values. Through this systematic calibration process, consistent and accurate radiance data is ensured across different spectral bands.

\subsubsection{Image Stitching}
\begin{figure}
\includegraphics[width=0.470\textwidth]{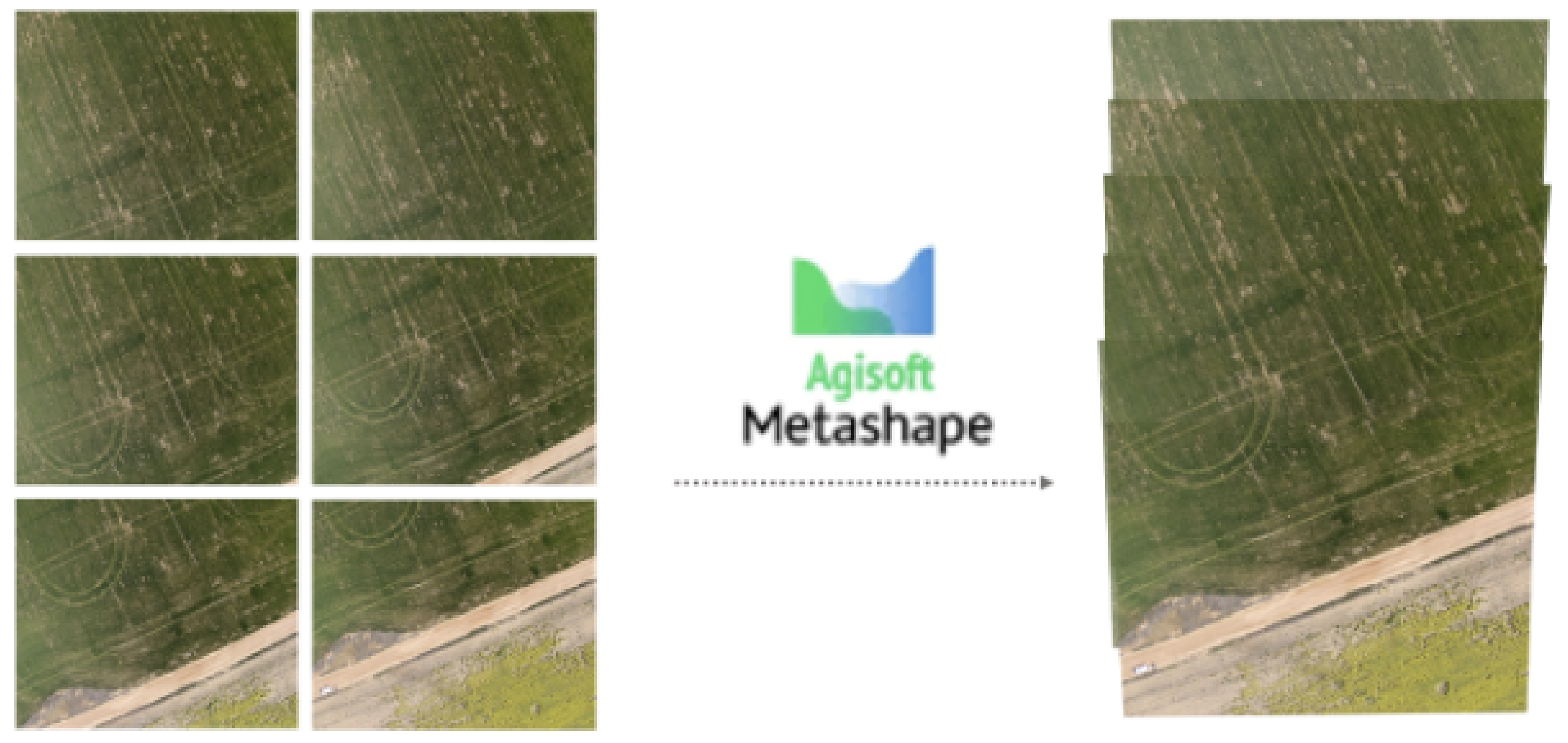} 
\vspace{-3mm}
  \caption{\footnotesize{Image stitching from raw data (RGB) to a map.}}
\label{agisoft}
\vspace{-6mm}
\end{figure}

To integrate multiple images into a single, continuous image covering the entire target area, Agisoft Metashape is used for panoramic stitching as shown in Fig.~\ref{agisoft}. Using the software's automated image processing capabilities, high-quality, seamless image stitching is achieved. The software automatically processes and merges the input multispectral images and parameter information, generating a seamless panoramic image. Multiple blending algorithms are applied to handle overlapping regions of the images, minimizing seams and stitching artifacts. It also completes orthographic correction. This ensures the creation of a continuous and uniform dataset.

\subsection{Feature Selection}
To ensure the accuracy of classification, it is essential to select appropriate features. The goal of feature selection is to extract and utilize the most representative information from the images to maximize the performance of the classification model. The multispectral dataset includes spectral features from several bands: Red, Green, Blue, NIR, RedEdge. 

Through extensive analysis of multispectral image data in agriculture and a thorough literature review\cite{vina2011comparison}, five key derived features are selected: Normalized Difference Vegetation Index (NDVI), Green Normalized Difference Vegetation Index (GNDVI), Enhanced Vegetation Index (EVI), Soil-Adjusted Vegetation Index (SAVI), and Modified Soil-Adjusted Vegetation Index (MSAVI). These features are widely used in agricultural remote sensing and have been proven to be highly valuable in crop and weed classification tasks \cite{yeom2019comparison}.

\textbf{Normalized Difference Vegetation Index (NDVI)}: NDVI is one of the most commonly used vegetation indices. It reflects the health and biomass of vegetation. High NDVI values typically indicate healthy vegetation, while low values suggest sparse or damaged vegetation \cite{yeom2019comparison}. NDVI is widely applied in agricultural monitoring. It is calculated through the following equation:

\begin{equation}
\label{eq:ndvi4}
NDVI = \frac{NIR - R}{NIR + R}.
\end{equation}

\textbf{Green Normalized Difference Vegetation Index (GNDVI)}:GNDVI is similar to NDVI but is more sensitive to changes in chlorophyll content in the vegetation. This makes it particularly effective in monitoring plant health and photosynthetic activity. GNDVI can provide more precise information on the health of vegetation, especially during the early stages of plant growth \cite{yeom2019comparison}. It is calculated through the following equation:

\begin{equation}
\label{eq:ndvi5}
GNDVI = \frac{NIR - G}{NIR + G}.
\end{equation}

\textbf{Enhanced Vegetation Index (EVI)}:EVI adjusts for atmospheric influences and soil background noise, providing more accurate vegetation information \cite{wendel2016self}. It is particularly useful in regions with dense vegetation cover. EVI complements NDVI by enhancing the vegetation signal in areas with high biomass. It is calculated through the following equation:

\begin{equation}
\label{eq:ndvi6}
EVI = 2.5 \times \frac{NIR - R}{NIR + 6 \times R - 7.5 \times B + 1}.
\end{equation}

\textbf{Soil-Adjusted Vegetation Index (SAVI)}:SAVI incorporates a soil brightness correction factor to minimize the impact of soil background on vegetation indices \cite{stroppiana2018early}. It is suitable for areas with significant soil interference. SAVI provides advantages in complex agricultural fields with significant soil background variation. It is calculated through the following equation:

\begin{equation}
\label{eq:ndvi7}
SAVI = \frac{NIR - R}{NIR + R + L} \times (1 + L),
\end{equation}
where $L$ is the soil adjustment factor, commonly set to 0.5.
\textbf{Modified Soil-Adjusted Vegetation Index (MSAVI)}:MSAVI is an improvement over SAVI, further adjusting the soil brightness factor for more stable vegetation monitoring results. MSAVI offers enhanced precision in areas with significant soil background changes \cite{huete2002overview}. It is calculated through the following equation:

\begin{equation}
\label{eq:msavi}
\scalebox{0.99}{$MSAVI = \frac{2 \times NIR + 1 - \sqrt{(2 \times NIR + 1)^2 - 8 \times (NIR - R)}}{2}$}.
\end{equation}

\begin{figure}
\includegraphics[width=0.470\textwidth]{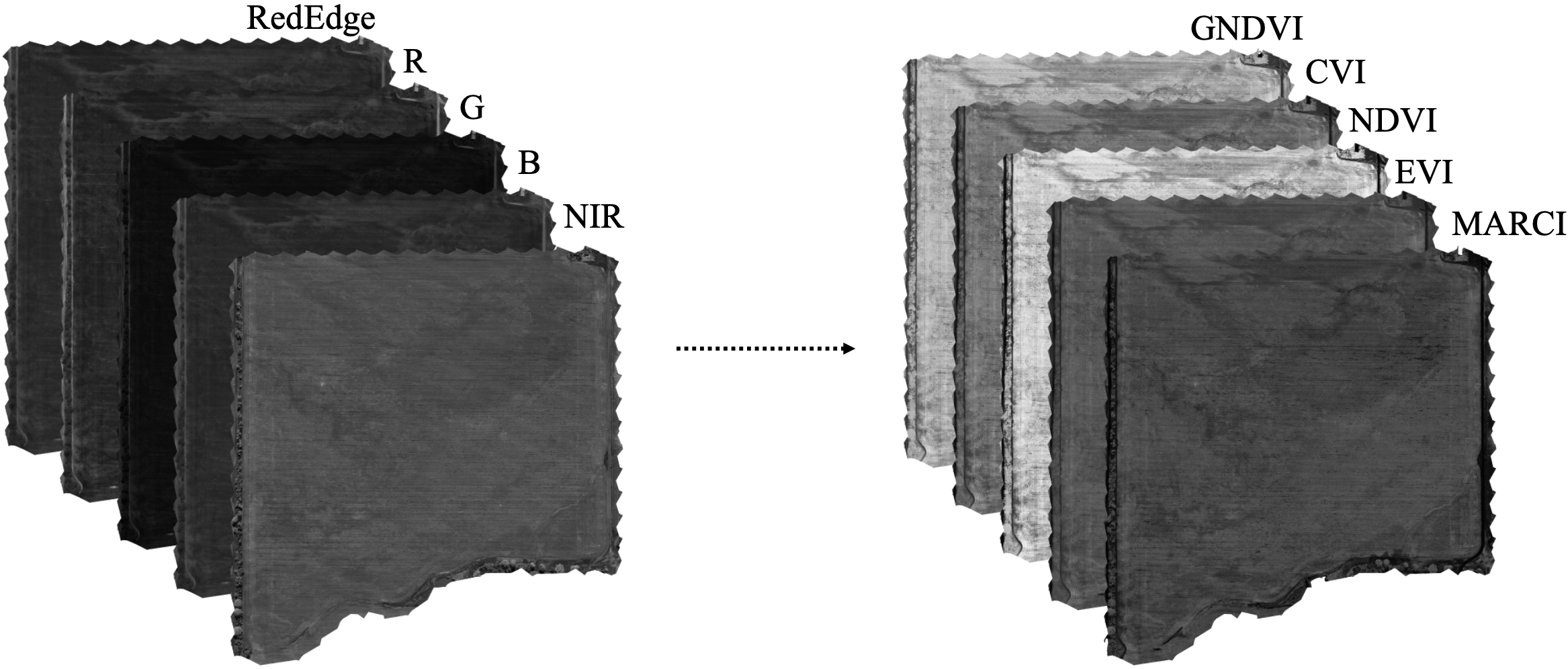} 
\vspace{-3mm}
  \caption{\footnotesize{5 selected features are calculated respectively using all 5 stitched multispectral images}}
\label{extraction}
\vspace{-6mm}
\end{figure}

The feature map is then calculated using all 5 original multispectral images as it is shown in Fig.~\ref{extraction}.

\subsection{Classification}
Five deep learning models are selected for weed and crop classification: ResNet-50\cite{bah2018deep}, U-Net\cite{ronneberger2015unet}, DeepLab\cite{chen2018deeplab}, SegNet\cite{badrinarayanan2017segnet}, and InceptionV3\cite{olsen2019deepweeds}. These models have demonstrated outstanding performance in remote sensing image analysis and object detection, each with unique architectural features that effectively process the collected multispectral data. 
 The whole process starts with data retrieval from the dataset, followed by data preprocessing and feature selection and extraction. The processed data is then divided into training, validation, and test sets. Multiple deep learning models are used for building the model, which are then evaluated. Based on the evaluation, model tuning is performed if necessary. Once the model is tuned and retrained, they are deployed and monitored. Result analysis is conducted to assess the performance of the deployed model.


ResNet, by introducing residual blocks, addresses the vanishing gradient problem in deep neural networks, allowing for deeper and more complex networks suitable for intricate remote sensing image classification task \cite{thomas2023weakly}. U-Net, a fully convolutional network designed specifically for biological image segmentation, has symmetrical downsampling and upsampling paths, enabling precise pixel-level classification \cite{ronneberger2015unet}. DeepLab employs dilated convolutions and Conditional Random Fields (CRF) for edge optimization, retaining high-resolution features while expanding the receptive field, thus achieving more accurate object segmentation \cite{xu2023instance}. SegNet, a deep learning model based on an encoder-decoder architecture, is particularly suited for semantic segmentation tasks, making it ideal for handling high-resolution remote sensing images \cite{subeesh2022deep}. InceptionV3, with its factorized convolutions and parallel convolution paths, reduces computational complexity while maintaining high accuracy, making it suitable for large-scale, diverse image datasets \cite{sa2018weedmap} \cite{sa2018weednet}. Each model offers distinct advantages, and together with our dataset and feature selection, they provide optimal classification performance across different application scenarios.

\section{Data Collection}
\label{sec:Data Collection}
In this section, the data collection process is detailed, including camera selection 
(which is DJI P4 Multispectral (P4M) Remote Piloted Aircraft
System (RPAS) and its multispectral camera), UAV flight path, camera calibration, and data organization. We collect the dataset used in this project spanning four years and totals 12,627 multispectral raw images with 543GB and created ground truth annotations by using GPS-enabled vehicles to manually label weeds and crops.

\noindent{\bf{Selection of UAV and Multispectral Camera}}: The DJI P4 Multispectral (P4M) Remote Piloted Aircraft System (RPAS) and its multispectral camera were opted for. The P4M system provides a stable flight platform, and its camera captures multiple bands including red, green, blue, near-infrared (NIR), and red edge (RedEdge), producing RGB, NIR, and RedEdge images.

\noindent{\bf{Flight Path Planning}}: To ensure comprehensive coverage of the target area and adequate overlap between images for later stitching, DJI Ground Station Pro (GSPro) is used for flight path planning \cite{dji2018manual}. The flight altitude was set at 120 meters, providing a large coverage area while maintaining high resolution for detailed analysis \cite{radoglou2020compilation}. Both forward and side overlaps were set at 80\% to ensure sufficient overlapping regions between images, enhancing the accuracy of image stitching and minimizing stitching errors \cite{bupathy2021optimizing}.
In our implementation, GSPro software was used to map the flight path. During flights, real-time monitoring of the UAV’s status and image capture was performed. All flight missions were conducted during times of direct sunlight under clear, cloudless weather conditions. Flights were executed strictly according to the pre-planned routes, ensuring stable UAV operation and avoiding abrupt changes in altitude or speed. The flight area covered 0.6046 square kilometers within two designated project experimental areas in Kondinin, Western Australia: E2 (lat: -32.508363, lon: 118.338139) and E8 (lat: -32.516563, lon: 118.353799).
\begin{figure}
\includegraphics[width=0.470\textwidth]{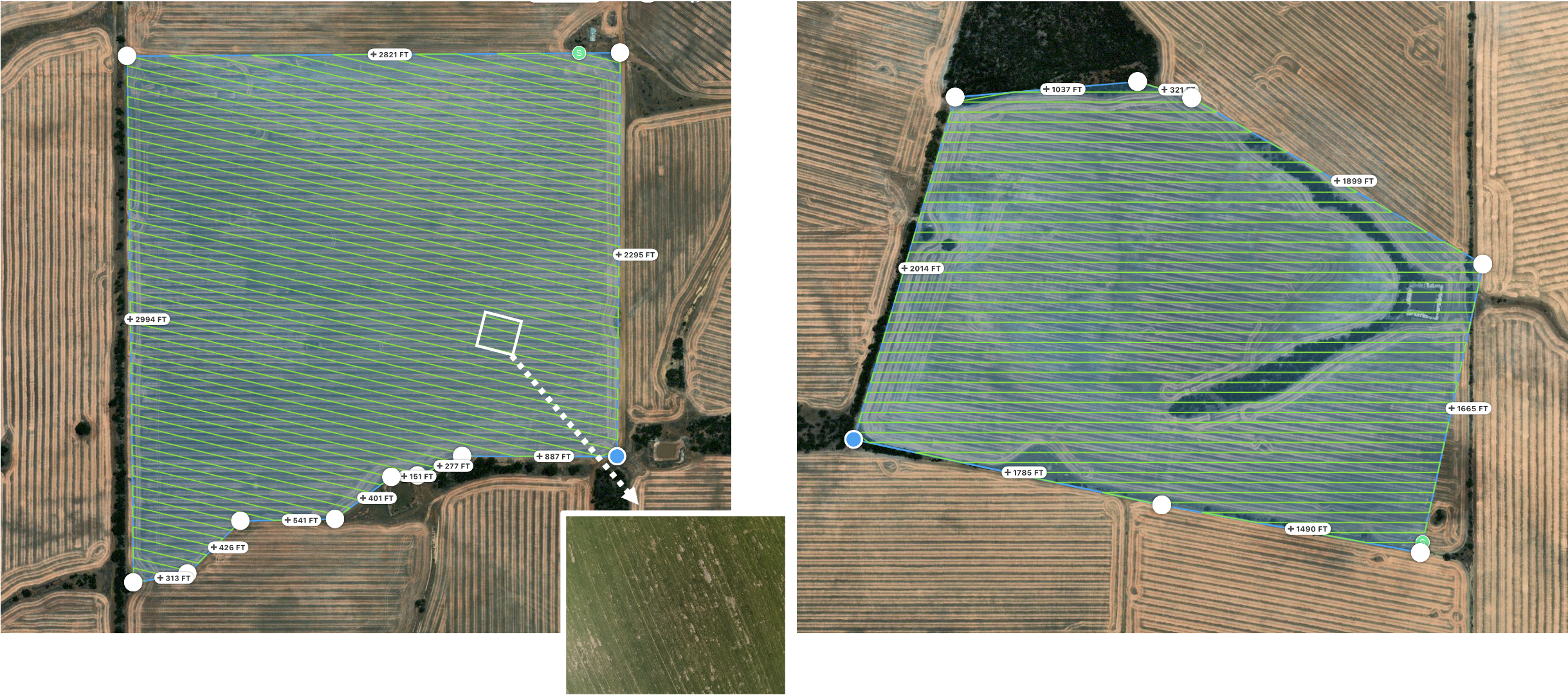} 
\vspace{-3mm}
  \caption{\footnotesize{Example (Left)Flight Path Planning for E2 experimental area(Right) Flight Path Planning for E8 experimental area with GSPro}}
\label{plan}
\vspace{-6mm}
\end{figure}

Fig.~\ref{plan} illustrates the planned flight paths for UAV image acquisition over 2 experimental fields. The left image (E8) demonstrates the designated flight path with specified waypoints to ensure comprehensive coverage, while the right image (E2) shows a different section of the target area with a similar approach. Throughout the flights, ground stations monitored the UAV’s flight status and image capture in real time, ensuring adherence to the pre-determined path and prompt issue correction. 

\noindent{\bf{Calibration of Multispectral Camera}}: Calibration of the multispectral camera is vital when conducting multiple flight missions at different times. The camera’s built-in calibration panel is used for reflectance calibration. Before each flight mission, the calibration panel is placed flat on the ground, unobstructed, and multiple images were captured for subsequent reflectance correction, ensuring consistency across different missions \cite{barker2019calibration}. Additionally, calibration panel images were captured under varying angles and lighting conditions to comprehensively account for environmental light changes \cite{barker2019calibration}. Environmental lighting conditions for each flight mission, including solar elevation angle, weather conditions (such as clear, partly cloudy, or overcast), temperature, and humidity, were meticulously recorded. These parameters were used in radiometric correction to adjust the image data, eliminating the effect from environmental changes.

\noindent{\bf{Data Organization and Recording}}: After each flight mission, image files are named according to a predefined naming convention, including the date of capture, mission number, and capture time (e.g., “20240423\_E2\_1230\_RGB.tif” for an RGB image captured in area E2 on April 23, 2024, at 12:30). Additionally, metadata for each flight mission, such as flight altitude (120.45 meters), flight speed, lighting conditions, and weather, are recorded as these are crucial for later data processing. Furthermore, the capture time and geographical coordinates of each image are recorded, ensuring precise geolocation during subsequent processing.

\begin{figure}[b!]
\includegraphics[width=0.470\textwidth]{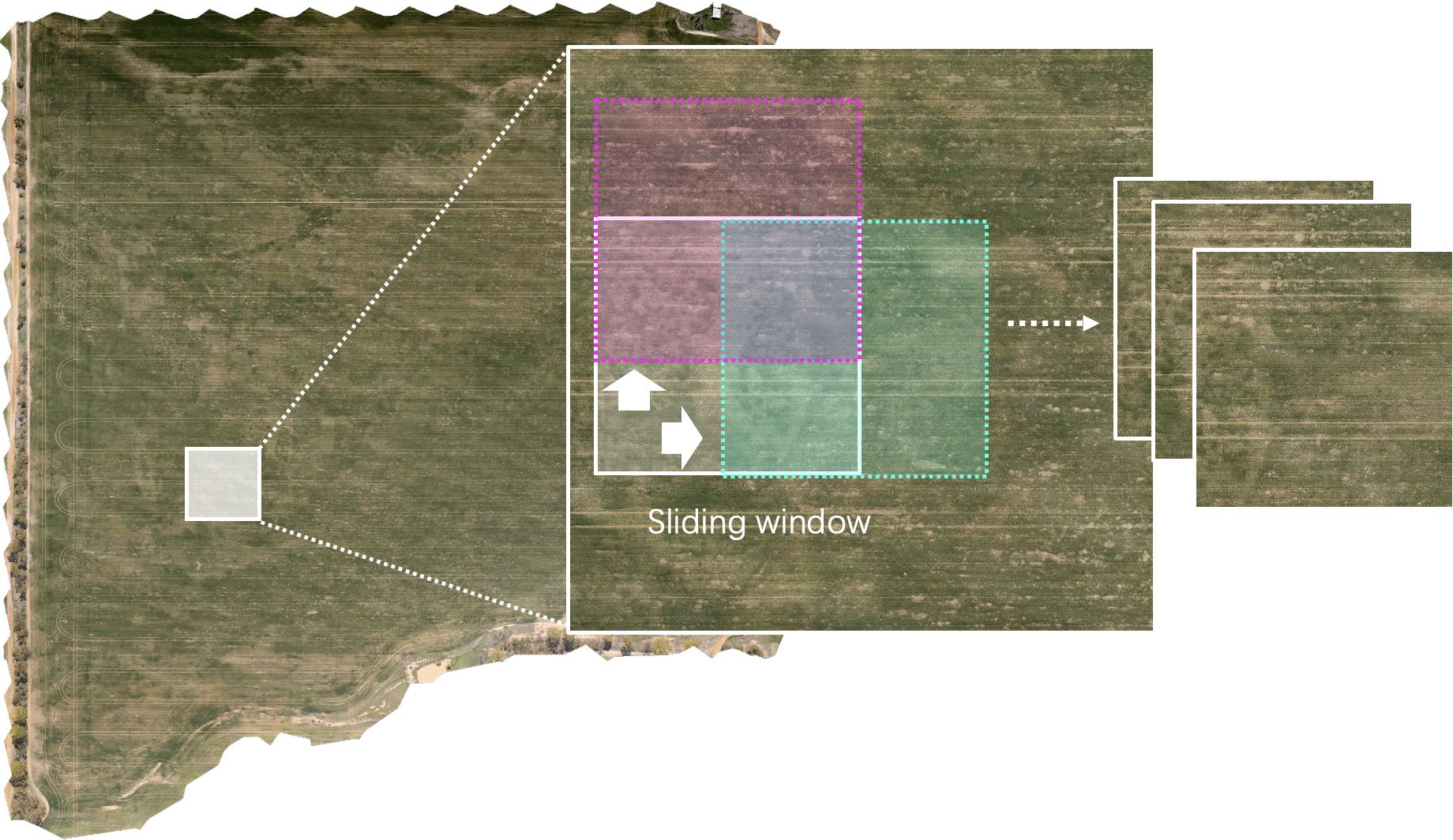} 
\vspace{-3mm}
  \caption{\footnotesize{An illustration demonstrating the application of the sliding window technique to systematically extract overlapping image patches from E2 image. The zoomed-in region, highlighted by the dashed white lines, focuses on a smaller section of the field. The arrow is the moving direction of the sliding window.}}
\label{sliding}
\vspace{-6mm}
\end{figure}

\section{Experiment}
\label{sec:Experiment}
In this section, we present a comprehensive evaluation of our proposed methodology for weed and crop classification using multispectral images. We describe the experimental setup and outline the steps involved in dataset formation, model training, and evaluation.  The performance of various deep learning models, including ResNet, U-Net, DeepLabV3+, InceptionV3, and SegNet, is compared across multiple metrics.  The results demonstrate the effectiveness of our approach and highlight the strengths and weaknesses of each model in different aspects of the classification task.

\subsection{Experiment Setup}
To carry out our experiment, we utilized a high-performance computing setup. The system configuration included an Intel Xeon E5-2670 v3 processor with 12 cores and 24 threads, operating at a base frequency of 2.3 GHz and a turbo frequency of 3.1 GHz. It is equipped with 64GB of HyperX DDR4 RAM. For graphics processing, two NVIDIA Tesla K40 GPUs are used, each with 12GB of GDDR5 memory and a computational power of 4.29 TeraFLOPS. The system ran on Ubuntu 18.04 LTS, ensuring a stable and efficient environment for our high-performance computing tasks.

\subsection{Dataset Formation}
To achieve efficient weed and crop classification, a training set is generated from the constructed multispectral dataset. Through image segmentation and data augmentation techniques, the diversity and robustness of the training set are ensured \cite{viola2001rapid}. Initially, the sliding window method is applied to segment the large multispectral map into multiple small patches. The image slice size was set to 512x512. Each patch size matches the input dimensions required by the deep neural network (DNN). Specifically, the sliding window moves across the large image at a fixed stride, generating a series of adjacent patches. The overlap between adjacent windows helps capture more edge information, thereby improving the robustness and continuity of the model. The sliding window method is illustrated in Fig.~\ref{sliding}.

Next, the generated patches are partitioned into subsets to ensure the model's generalization capability and the reliability of performance evaluation. The data are divided as follows: 70\% for training, 10\% for validation, and 20\% for testing. The training set is used to train the model. The validation set is used during training to evaluate model performance and adjust hyperparameters, thereby avoiding overfitting. The test set evaluates the model's performance on unseen data.

\begin{figure}
\includegraphics[width=0.470\textwidth]{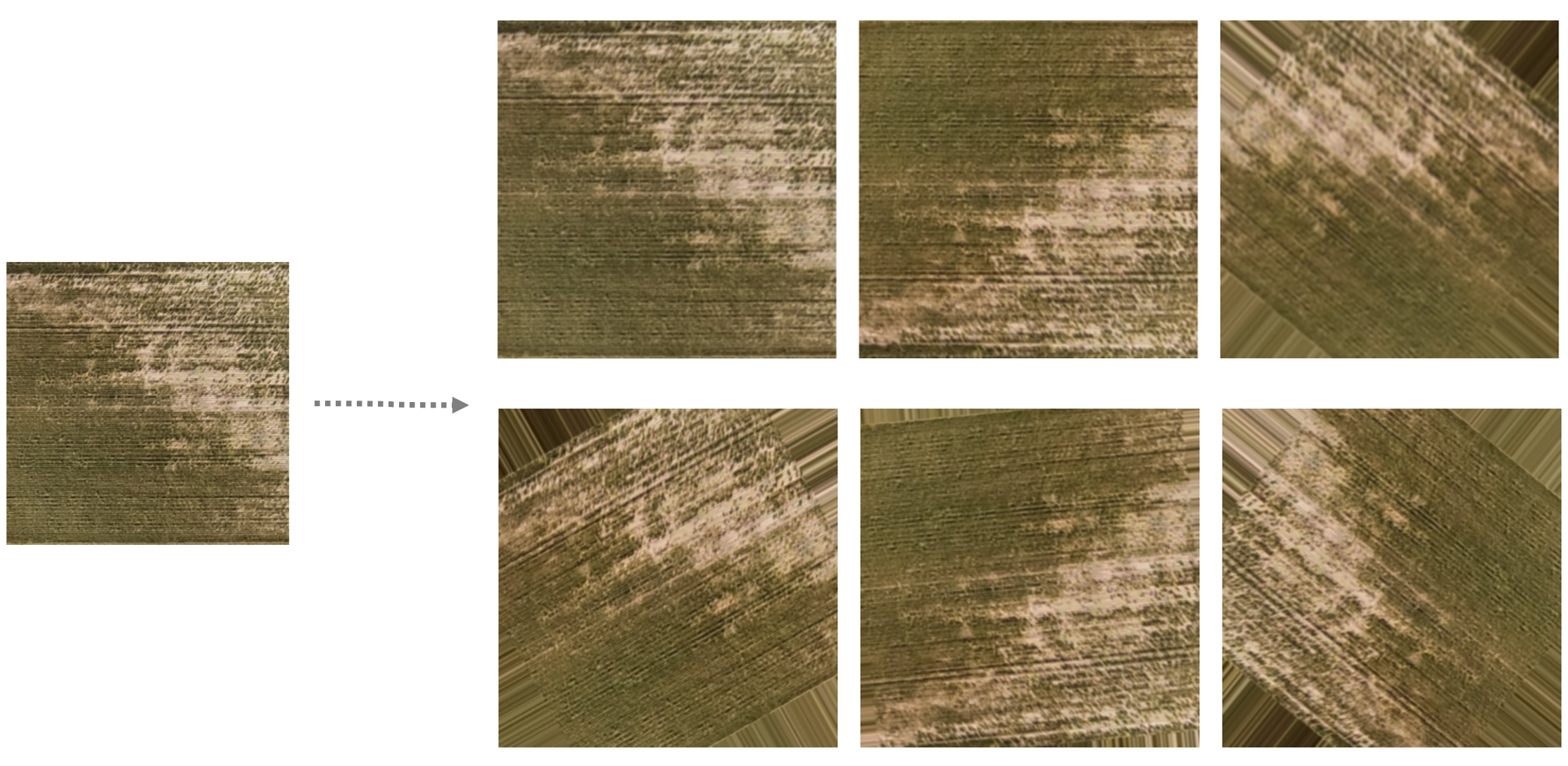} 
\vspace{-3mm}
  \caption{\footnotesize{Example of Image Augmentation for training set (R, G, B spectra combined): left is the original image, by image augmentation, 6 variants will be generated and added to the training set to make the model keep robust throughout different scales and orientations.}}
\label{augmentation}
\vspace{-6mm}
\end{figure}

\begin{table*}[htbp]

\caption{\footnotesize{Results of our proposed end-to-end framework for weed detection using ResNet, U-Net, DeepLabV3+, InceptionV3, SegNet on our proposed dataset. The metrics including accuracy, mean F1-Score among 3 classes, mIOU and mDC are used for model evaluation}}

\label{tab:comparison}
\begin{center}
\begin{tabular}{p{3cm} p{3cm} p{3cm} p{3cm} p{3cm} }
\hline
\textbf{Model} & \textbf{Accuracy} & \textbf{F1-Score (mean)} & \textbf{mIOU} & \textbf{mDC} \\
\hline
ResNet & 0.9213 & 0.873488 & 0.7888 & 0.88648783 \\
U-Net & 0.80488135 & 0.74236548 & 0.69636628 & 0.77840223 \\
DeepLabV3+ & 0.82151894 & 0.76458941 & 0.63834415 & 0.79627983 \\
InceptionV3 & 0.81027634 & 0.74375872 & 0.6791725 & 0.7535518 \\
SegNet & 0.80448832 & 0.7891773 & 0.65288949 & 0.75435646 \\
\hline
\end{tabular}
\end{center}
\end{table*}

To further enhance the model's robustness and generalization ability, various data augmentation techniques are applied to the training data. These methods include random rotation, flipping, and Gaussian blur \cite{simard2003best}. Fig.~\ref{augmentation} is an example of how the training set is formed through image augmentation. 

These data augmentation techniques generate additional training samples, increase data diversity, and simulate different image qualities and shooting conditions, ensuring the model's adaptability to various scenarios and complex environments. By segmenting the large image using the sliding window method, applying reasonable data partitioning, and employing multiple data augmentation techniques, a high-quality training set is generated. This provides a solid data foundation for the subsequent weed and crop classification model, ensuring the model's efficiency and reliability in practical applications.


\vspace{-3mm}

\begin{figure}[htbp]
\includegraphics[width=0.470\textwidth]{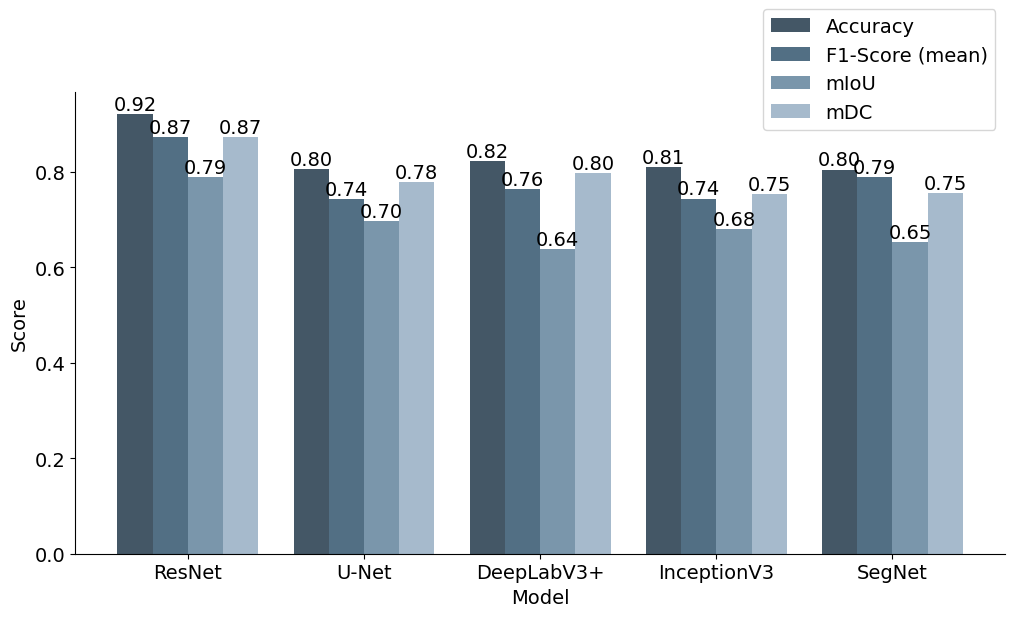} 
\vspace{-3mm}
  \caption{\footnotesize{Performance Comparison of Different Deep Learning Models in Weed Corp classification task in test set of E2 experimental area. Each colour represents one metric.}}
\label{bar}
\vspace{-6mm}
\end{figure}

\subsection{Model Evaluation}
Four key metrics are used to evaluate the performance of these models: Precision, F1-score, Mean Intersection over Union (mIOU), and Mean Dice Coefficient (mDC). These metrics provide a comprehensive assessment of the models' performance in classification tasks. Precision measures the accuracy of the models' predictions, focusing on the false positive rate when identifying weeds and crops. F1-score combines precision and recall, offering a balanced performance evaluation, particularly useful for datasets with class imbalances. mIOU is a crucial metric for image segmentation tasks, assessing the overlap between predicted segmentation and ground truth\cite{long2015fully}. The Mean Dice Coefficient (mDC) reflects the similarity between the segmentation results and the ground truth, sensitive to boundary details, ensuring that the models perform well in handling complex boundaries \cite{milletari2016vnet}. By utilizing these metrics, the models' performance can be thoroughly evaluated across multiple dimensions, ensuring efficient weed and crop classification in practical applications. 



We selected ResNet-50 \cite{bah2018deep}, U-Net \cite{ronneberger2015unet}, DeepLabV3+\cite{chen2018deeplab}, InceptionV3\cite{olsen2019deepweeds}, and SegNet\cite{badrinarayanan2017segnet} due to their proven effectiveness in handling complex remote sensing image data \cite{bah2018deep, olsen2019deepweeds}, with each model offering unique strengths in feature extraction, pixel-level segmentation, and computational efficiency \cite{bah2018deep, badrinarayanan2017segnet}, allowing for a comprehensive evaluation across different aspects of the weed detection task. The selected models are applied to our formatted dataset for weed and crop classification. Their performances are then compared across the four evaluation metrics. Our analysis revealed that each model has strengths and weaknesses in different metrics. The metrics for each model is displayed in Table~\ref{tab:comparison}.

Fig.~\ref{bar} is the bar chart which provide a more intuitive comparison of all 5 models. ResNet excelled in overall performance, particularly in Precision and F1-score, indicating its high reliability in accurately identifying weeds and crops. {ResNet-50’s residual connections not only resolves the vanishing gradient problem but also allows for deep feature extraction, enabling fine-grained pixel-level analysis of differences between weeds and crops. Particularly in multispectral images, the deep network of ResNet captures subtle variations across spectral bands, improving pixel-level classification accuracy\cite{bah2018deep}. U-Net demonstrated significant advantages in handling high-resolution images, with superior detail processing. DeepLab, with its powerful feature extraction and edge optimization capabilities, performed exceptionally well in mIOU and mDC, showing its advantage in overall segmentation quality. SegNet exhibited good performance in computational efficiency and segmentation details, making it suitable for large-scale datasets requiring efficient processing. 

\begin{figure}
\includegraphics[width=0.470\textwidth]{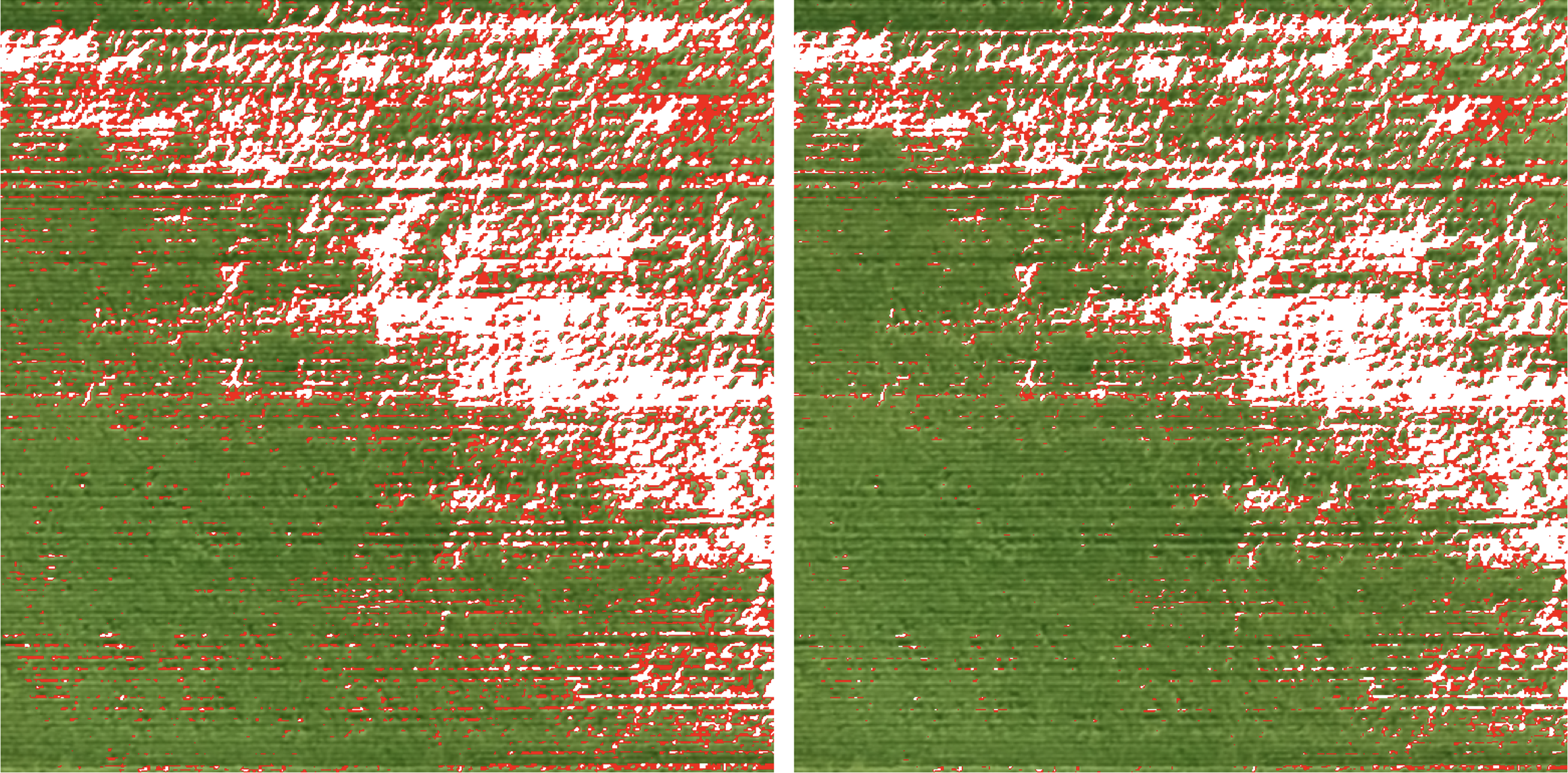} 
\vspace{-3mm}
  \caption{\footnotesize{After applying ResNet to obtain the classification results, crops are mapped in green, weeds are mapped in red, and the background is mapped in white. (Left) Ground truth. (Right) Predicted result.}}
\label{pred}
\vspace{-6mm}
\end{figure}
Through detailed comparison and analysis of these models, the most suitable model which is ResNet is selected for practical application needs, achieving efficient and accurate weed and crop classification. The result is shown in Fig.~\ref{pred}

\vspace{-2mm}
\section{Conclusion}
\label{sec:Conclusion}
This study addresses the agricultural challenges in Western Australia's Kondinin region caused by pervasive weed infestations. We developed a tailored multispectral remote sensing dataset using the DJI Matrice 300 RTK UAV. Over four years, data were collected from two experimental areas (E2 and E8), covering 0.6046 km². Comprehensive preprocessing, including denoising, radiometric calibration, and stitching, produced a high-resolution labeled dataset. Key vegetation indices (NDVI, GNDVI, EVI, SAVI, MSAVI) were selected as features. We proposed an end-to-end framework for weed detection, integrating data preprocessing, feature selection, deep learning model training, and prediction generation. ResNet achieved the highest performance metrics, validating the dataset's efficacy. The primary contributions are the construction of a multispectral remote sensing dataset specifically designed for Western Australian agriculture and the proposal of an end-to-end framework for weed detection based on this dataset.

\vspace{-3mm}
\section*{Acknowledgment}
\vspace{-2mm}
This research is supported by the Department of Primary Industry and Regional Development (DPIRD) and the University of Western Australia. Professor Ajmal Mian is the recipient of an Australian Research Council Future Fellowship Award (project number FT210100268) funded by the Australian Government.

\vspace{-1mm}
\bibliographystyle{IEEEtran}
\bibliography{DICTAsubmision}

\end{document}